# Exploring Automated Essay Scoring for Nonnative English Speakers

Amber Nigam

Independent Scholar, New Delhi, India
ambernigam12@gmail.com

**Abstract.** Automated Essay Scoring (AES) has been quite popular and is being widely used. However, lack of appropriate methodology for rating nonnative English speakers' essays has meant a lopsided advancement in this field. In this paper, we report initial results of our experiments with nonnative AES that learns from manual evaluation of nonnative essays. For this purpose, we conducted an exercise in which essays written by nonnative English speakers in test environment were rated both manually and by the automated system designed for the experiment. In the process, we experimented with a few features to learn about nonnative phraseology and its impact on the manual evaluation of the essays. The proposed methodology of automated essay evaluation has yielded a correlation coefficient of 0.750 with the manual evaluation.

## 1   Introduction

There are different versions of Automated Essay Scoring (AES) and lack of generalizability across different analyses and corpora prompts a question over the validity of one-size-fits-all AES.

Furthermore, nonnative analysis is differentiated from the native analysis on a few aspects. For example, it is difficult to detect context in essays that have errors specific to some nonnative usages. Moreover, a few valid nonnative spellings like Qutub Minar and Karur are not a part of standard English. This, however, does not render the usage of these words incorrect.

In this paper, we have discussed our methodology for nonnative essay evaluation. First, we have described the feature set that we used for our experiments. Second, we have discussed various adjustments that were made to the system to make it learn and account for nonnative phraseology from manual evaluation. Finally, we have discussed the results of our experiments.

## 2   Related Work

Although Automated Essay Scoring (AES) has been widely used in many of the real-world applications, there is very limited published work on rating nonnative speakers. Following analyses deal with nonnative speakers in one way or another.

e-rater system™, developed by Educational Testing Service (ETS), is one of the tools that automates scoring of English essays of native and nonnative speakers. In their analysis of e-rater, Jill Burstein, et al. (1999), reported that even when 75% of essays used for model building were written by nonnative English speakers, the features selected by the regression procedure were largely the same as those in models based on operational writing samples in which most of the sample were native English speakers. The correlations between e-rater scores and those of a single human reader were about .73. However, as mentioned in the paper, there were significant differences between final human reader score and e-rater score across language groups, and more data is needed to build individual models for different language groups to examine how this affects e-rater's performance.

Another analysis in this field is by Sowmya Vajjala, (2016), which is the first multi-corpus study using TOEFL11SUBSET and First Certificate in English (FCE) datasets. For TOEFL11SUBSET, the best model achieved a prediction accuracy of 73% for classifying between three proficiencies (low, medium and high), using all the features. In general, shallow discourse features such as word overlap were more predictive for this dataset compared to deeper ones like reference chains. The native language of the author was an important predictor for some feature groups in this dataset. With FCE, the best model achieved a correlation of 0.64 and a Mean Absolute Error of 3.6, on the test data. In general, features that relied on deeper linguistic modeling (such as reference chains) had more weight for this dataset compared to other features. The native language of the author was not an important predictor in the dataset. The study concludes by stating that the features do not seem to be completely generalizable across datasets.

Current research differs from the existing research in its feature set, unique methodology, and an essay corpus composed of essays written by candidates whose native language is Hindi. Besides, some of our features like lexical density and readability have either not been reported or reported with less significance in earlier analyses on nonnative AES. We have also successfully implemented grammar error correction for a better context detection (Alla Rozovskaya and Dan Roth, 2016) and an automated correction mechanism for whitelisting nonnative spellings that are not a part of standard English.

We are not referring to generic AES systems like by Dimitrios Alikaniotis, et al. (2016) and by Kaveh Taghipour, et al. (2016) because current paper is only concerned with nonnative AES.

## 3   Experiment

The experiment encompassed building an automated scoring model after learning from training essays (as shown in Figure 1). In all, there were more than 900 essays of length between 150 and 400 words across 7 unique topics (see Table 1). The mean and the standard deviation for word count of these essays were 247 and 43 respectively. Essay topics were carefully chosen to be on commonly known issues so that they are easy to comprehend and write about. The test takers were all undergraduate students whose native language is Hindi to keep the analysis independent of test taker's native language. Each essay was manually scored on a scale of 1-10 by two raters, both Professors of English at a reputable Indian University, and they had .81 Cohen's kappa statistic (Cohen J, 1968) between their ratings. The raters were informed about the overall experiment and its intent, which is to holistically judge written English of the essays majorly on content, coherence, complexity, and adherence to rules.

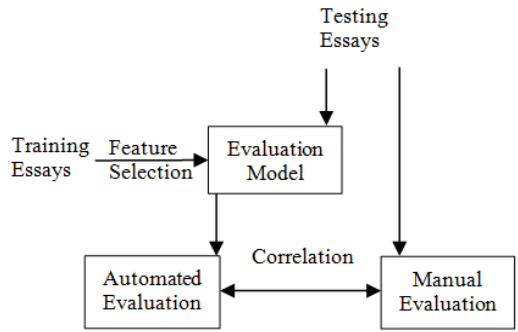

Figure 1: Process flow of the experiment

We used LanguageTool (Daniel Naber, 2003; LanguageTool, 2012) for detecting grammatical errors. Other features were evaluated by the software developed for the experiment. Besides, the software was designed to detect cheating attempts such as repeating content, writing out of context, and excessive usage of irrelevant words. Feature selection and machine learning experiments were done using Waikato Environment for Knowledge Analysis (Weka) toolkit (Hall et al., 2009). For machine learning experiments, the split between training, validation, and testing sets was approximately 60:20:20.

| Essay Topics |
| --- |
| Corruption in politics |
| Role of ambition in career |
| Factors of motivation for an employee |
| Power leads to Corruption |
| Importance of leadership qualities |
| Should juveniles be tried as adults? |
| Adverse effects of climate change |

Table 1: Essay topics used in the experiment

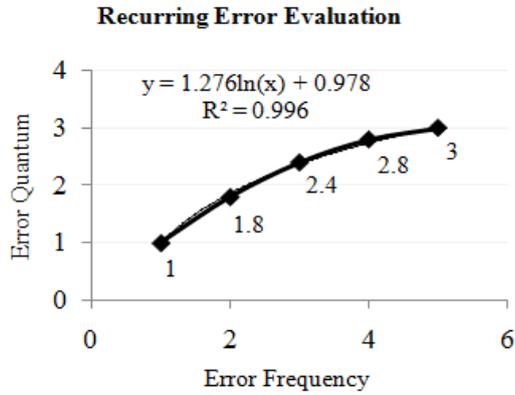

Figure 2: Variation of error quantum with repeated error frequency for a single word

## 4 Feature Set Selection

| Feature | Correlation |
|---|---|
| Grammar Error Density | -0.396 |
| Grammar Error Coverage | -0.295 |
| Spelling Error Density | -0.146 |
| Spelling Error Coverage | -0.137 |
| Readability | 0.326 |
| Lexical Density | 0.393 |

Table 2: Prominent features and their correlations with the manual score

We used Principal Component Analysis (PCA) to choose our final feature set from a larger set of features. Correlations between manual scores and the top few features were then evaluated (see Table 2) to understand the relative significance of the features in manual evaluation. The top features are described below:

### 4.1 Grammar error density (i.e. grammar errors per unit length)

In this paper, grammar error density refers to the grammatical error count per 100 words. We use density rather than simple error count to ensure that candidates who have written longer essays are not unduly penalized over those who have written shorter essays with the same grammar error density.

A granular categorization of grammar errors has helped in classifying the errors into severity bands based on their impact on manual evaluation. For example, it was observed that a subject-verb agreement error was generally more severely penalized than a style based error in the manual evaluation. Following are the buckets in which grammar errors have been classified:

**Major errors like wrong form, incorrect tense, and agreement errors:** This bucket includes severe grammar errors that degrade the quality of the essay and possibly cause serious comprehension issues. For instance, replacing "his" with "her" to form a sentence like "He is known for her intelligence." would be detrimental to the semantics of the sentence.

**Capitalization errors:** Such errors occur when case of the character is not what it ought to be. This might happen when the first word of a sentence begins with a letter in lowercase. Proper nouns like Paris, Shakespeare also need to be capitalized. Also, personal pronoun "I" is always written in capital when used. Other pronouns are capitalized only if they begin a sentence.

**Typography errors:** These are also known as typographical errors and are not same as spelling errors. They result due to mechanical failure or slips of the hand or finger. For instance, "water" typed as "wster" due to S key being close to the A key. Typos generally involve duplication, omission, transposition or substitution of a small number of characters.

**Style based errors:** These errors include usage of informal language or shorthand like using "u" instead of "you".

**Common replacement errors:** Such errors happen in case of a sound-alike or looka-like word pair when one of the words is replaced by the other word in the pair. An example of such word pair is "affect" and "effect".

**Punctuation errors:** These errors refer to incorrect usage of comma, semi-colon, colon, apostrophe, hyphen, etc. in a sentence or paragraph. Misplaced punctuation can sometimes alter the meaning of a sentence. For example, a sentence like "Jane finds inspiration in cooking, her family, and her dog." without commas would be read as "Jane finds inspiration in cooking her family and her dog.".

**Miscellaneous errors:** All other grammar errors are put under miscellaneous bucket. These include errors such as repetition of words, improper white space usage, etc.

## 4.2  Grammar error coverage

It is the count of type of grammar errors per 100 words in an essay. Grammar error coverage highlights the spread of errors across different grammar buckets.

## 4.3  Spelling error density (i.e. spelling errors per unit length)

Spelling error density of an essay is referred to as the number of spelling errors in the essay per 100 words.

Penalty for a recurring error is based on error quantum, which is evaluated by damping the error frequency (as shown in Figure 2), to lower incremental penalty for a single recurring error.

### 4.4 Spelling error coverage

It is the count of unique spelling errors per 100 words in an essay. The feature compliments spelling error density by accounting for cases where a difficult spelling might be repeated in the essay due to a difficult topic.

### 4.5 Readability

Readability attempts to estimate the complexity of phraseology used in a text. Why is it relevant in nonnative speakers' analysis?

It helps distinguish between the candidates who can articulate their thoughts into a syntactically correct and, if required, complex structure, and those who cannot. Because some nonnative speakers lack even the basic ability of constructing appropriate text, it becomes a very important feature. Our data shows that readability alone has a strong correlation with manual scoring. We use Flesch–Kincaid grade level (Kincaid JP, 1975) that is evaluated on word count (WC), sentence count (SC), and syllable count (SyC) by the equation:

$$0.39 * \left(\frac{WC}{SC}\right) + 11.8 * \left(\frac{SyC}{WC}\right) - 15.59 \qquad (1)$$

### 4.6 Lexical Density

It measures the ratio of lexical words to total words that include lexical and grammatical words (Ure, J 1971). Lexical words give a text its meaning and include nouns, adjectives, most verbs, and most adverbs. Grammatical words act as syntactic sugar and include pronouns, prepositions, conjunctions, etc. Lexical density gives a measure of the breadth of content in an essay. This is another factor that is strongly correlated with manual scoring.

The formula for lexical density (LD) is:

$$\frac{N_{lex}}{N} * 100 \qquad (2)$$

where $N_{lex}$ is number of lexical word tokens (nouns, adjectives, verbs, adverbs) and N is the total number of tokens in the text.

### 4.7 Context/Relevance

Latent semantic analysis (Foltz, et al., 1998) is used to detect the context of the topic using n-grams of phrases from word count 1 to 5. Our data shows that certain nonnative erroneous multi-word expressions (MWE) make context detection difficult and that we are better able to detect context in the essays with such errors through grammatical error correction (Alla Rozovskaya and Dan Roth, 2016). Context is detected by checking if cosine similarity between vector of the evaluated essay and vector of at least one corpus essay is more than a specified threshold.

| Algorithm | Correlation |
|---|---|
| Random Forest | 0.750 |
| Random Subspace | 0.738 |
| Bagging | 0.731 |
| M5 Rules | 0.706 |
| Gaussian Processes | 0.681 |

Table 4: Correlations between scores by machine learning algorithms and manual scores

## 4.8 Coherence

We use Grid model (Lapata and Barzilay, 2005; Barzilay and Lapata, 2008) to detect coherence. This attribute accounts for the flow and structuring of an essay.

## 5 Self-correction Mechanism

One of the unique selling propositions of the engine is that it pops unknown spellings/phrases once their cumulative frequency is beyond a pre-defined threshold. This has helped us add many nonnative spellings into our repository.

## 6 Results

We used machine learning algorithms to predict the scores of essays and Random Forest algorithm's predictions were closest to the manual scoring as shown in the Table 3. The correlations that we report are statistically significant (given the parameters of the experiment) for a significance level of 0.01. Intuitively, error densities like grammatical error density and spelling error density that predict adherence to rules are among the top predictive features. Besides, lexical density and readability are also some of the important features, which underscores the importance of a lexical complexity metric for nonnative Automated Essay Scoring (AES).

## 7 Conclusion

In this paper, we presented a methodology of rating English essays of nonnative speakers that includes but is not limited to selecting a relevant feature set for the evaluation, categorizing grammar errors into finer types to learn about their importance from their distinctive treatment in contribution to the manual evaluation in nonnative context, treating essays for typical nonnative grammar errors (Alla Rozovskaya and Dan Roth, 2016) that improved context matching for essays with nonnative errors, and devising a self-correction mechanism to learn and promptly address nonnative spellings and styles. Our results show that these incremental adjustments have cumulatively helped in better alignment of automated evaluation with manual evaluation.

# References

Alla Rozovskaya and Dan Roth. Grammatical Error Correction: Machine Translation and Classifiers ACL (2016)

Cohen, J.: Weighted Kappa: Nominal Scale Agreement with Provision for Scaled Disagreement or Partial Credit. Psychol. Bull. 70, 213–220 (1968)

Daniel Naber. A Rule-Based Style and Grammar Checker, Diploma Thesis, University of Bielefeld, 2003

Dimitrios Alikaniotis, Helen Yannakoudakis, Marek Rei. Automatic Text Scoring Using Neural Networks. In Proceedings of ACL, pp. 715–725, 2016.

Kaveh Taghipour, Hwee Tou Ng. A Neural Approach to Automated Essay Scoring. In Proceedings of EMNLP, pp. 1882–1891, 2016.

Foltz, P. W., Kintsch, W. & Landauer, T. K. (1998). The measurement of textual coherence with Latent Semantic Analysis. Discourse Processes, 25(2&3), 285-307.

Hall, M., Frank, E., Holmes, G., Pfahringer, B., Reutemann, P., and Witten, I. H. (2009). The weka data mining software: An update. In The SIGKDD Explorations, volume 11, pages 10–18.

Jill Burstein and Martin Chodorow. Automated Essay Scoring for Nonnative English Speakers

Kincaid JP, Fishburne RP, Rogers RL, Chissom BS. Derivation of New Readability Formulas (Automated Readability Index, Fog Count and Flesch Reading Ease Formula) for Navy Enlisted Personnel. Memphis, Tenn: Naval Air Station; 1975.

LanguageTool. Style and Grammar Checker. Retrieved 2013-09-04 from http://www.language-tool.org/; 2012

Mirella Lapata and Regina Barzilay. 2005. Automatic evaluation of text coherence: Models and representations. In IJCAI, volume 5, pages 1085–1090.

Regina Barzilay and Mirella Lapata. 2008. Modeling local coherence: An entity-based approach. Computational Linguistics, 34(1):1–34.

Sowmya Vajjala. Automated assessment of non-native learner essays: Investigating the role of linguistic features

Ure, J (1971). Lexical density and register differentiation. In G. Perren and J.L.M. Trim (eds), Applications of Linguistics, London: Cambridge University Press. 443-452.